\documentclass[a4paper]{article}

\usepackage{INTERSPEECH_v2}

\title{Weakly Supervised PLDA Training}
\name{Lantian Li$^1$, Yixiang Chen$^1$, Chenghui Zhao$^2$, Dong Wang$^1$$^*$}
\address{
  $^1$Center for Speech and Language Technologies, Tsinghua University, China \\
  $^2$Pachira, China}
\email{\{lilt13,wangdong99\}@mails.tsinghua.edu.cn, chenxy@cslt.riit.tsinghua.edu.cn \\
zhaochenghui@pachiratech.com
}

\begin{document}

\maketitle
\begin{abstract}

  PLDA is a popular normalization approach for the i-vector model,
  and it has delivered state-of-the-art performance in speaker verification.
  However, PLDA training requires a large amount of labelled development
  data, which is highly expensive in most cases. We present a cheap
  PLDA training approach, which assumes that speakers in the same session
  can be easily separated, and speakers in different sessions are simply different. This results in `weak labels' which are
  not fully accurate but cheap, leading to a weak PLDA training.

  Our experimental results on real-life large-scale telephony customer service demonstrated that this weak PLDA training can offer good performance
  when human-labelled data are limited.
  More interestingly, the weak training can be employed as
  an adaptation approach, which is more efficient than the prevailing unsupervised method
  when human-labelled data are insufficient.

\end{abstract}

\section{Introduction}
\label{sec:intro}

The i-vector model plus various normalization approaches offers the standard framework
for modern speaker verification~\cite{dehak2011front,garcia2011analysis,lei2014novel,kenny2010bayesian}.
Basically, the i-vector model uses a Gaussian mixture model (GMM) or a deep neural network (DNN)
to collect the Baum-Welch statistics, based on which an affine transform is learned so that
speech segments can be projected onto low-dimensional continuous vectors (i-vectors).
Although it is possible to discriminate speaker i-vectors using simple
cosine distance, normalization or discriminative techniques are often preferred, since they promote speaker-related information and thus bring significant performance improvement. Probabilistic
linear discriminant analysis (PLDA) is one of the most popular normalization methods.
It assumes that i-vectors of a particular speaker are subject to a Gaussian
distribution, with the mean vector following a normal distribution~\cite{garcia2011analysis}.
Combined with length normalization, PLDA has delivered state-of-the-art performance in
various test benchmarks~\cite{kenny2010bayesian}.

PLDA training generally requires a large amount of human-labelled data, usually thousands of speakers,
each with multiple sessions.
For example, in the two popular development databases Fisher~\cite{fisher2004} and Switchboard~\cite{switch},
there are $12,399$ and $543$ speakers, respectively.
In practice, labelling such a large amount of data by human is very challenging: it is
not only because discriminating two voice-similar speakers is difficult, but also because
identifying the speaker of an utterance among thousands of people is nearly impossible.
Therefore, it is quite appealing if the data can be utilized directly
without human labeling.

A popular approach towards this direction is various unsupervised adaptation techniques.
For example, Wang et al.~\cite{wang2016domain} proposed a domain-adaptation
approach based on maximum likelihood
linear transformation (MLLT), and Rahman et al.~\cite{rahman2015dataset} proposed a dataset-invariant
covariance normalization approach that normalized i-vectors by a global covariance matrix
computed from both in-domain and out-domain data. This is equal to projecting i-vectors of in-domain
and out-domain speakers onto a dataset-invariant space, so that the PLDA model trained with
the projected i-vectors is more robust against data mismatch.

Another approach to utilizing unlabelled data is to produce labels for these data automatically. These
labels may be not as accurate as human labels but still convey some speaker-related information, and
therefore can be used as supplemental materials in PLDA training. Most importantly, these labels are
very cheap, allowing vast unlabelled data to be used. We call these cheap labels `weak labels',
and the PLDA training based on these labels `weak training'. Correspondingly, the PLDA training with
human labels is called `strong training'.

Some research has been conducted on weak PLDA training.
Garcia-Romero et al.~\cite{garcia2014improving} proposed a semi-supervised learning approach that
used an out-of-domain PLDA to cluster in-domain
data, based on which the PLDA projection matrix was adapted.
Villalba and colleagues~\cite{villalba2014unsupervised} proposed a variational Bayesian method
where the unknown label of an unlabelled utterance was treated as a latent variable. This can be seen
as an extension of the semi-supervised method.
Liu et al.~\cite{liu2014utilization} proposed an approach that treated unlabelled data
as from a special universal speaker, and the PLDA was trained with the universal speaker involved.

This paper proposes a new knowledge-based weak PLDA training approach that produces cheap labels based
on some prior knowledge.
For example, in the telephony customer service domain,
the prior knowledge is that speakers in different sessions are almost different,
and therefore the session ID can be used to label speakers. These labels
are certainly noisy (therefore weak) since the knowledge is not absolutely
correct, but they do convey some valuable information that can be used to
enhance PLDA. Our experiments on a real-life large-scale customer service archive
demonstrated that the knowledge-based weak training is rather effective in domains
where the knowledge is `sufficiently correct' and can provide performance
improvement, and even outperform the unsupervised adaptation approach in scenarios when
human-labelled data are limited.

The structure of this paper is as follows: Section~\ref{sec:local} presents details of weak training,
and Section~\ref{sec:exp} presents the experiments. Finally Section~\ref{sec:conc} concludes the
paper and discusses some future work.

\section{Knowledge-based weak PLDA training}
\label{sec:local}

In this section, the conventional PLDA model is briefly reviewed, and then our proposed knowledge-based weak
training approach is presented. We also discuss the relation of our proposal methods and some others.

\subsection{PLDA model}

PLDA is an extension of the linear discriminative analysis (LDA), by introducing a Gaussian
prior on the mean i-vector of each speaker. Combined with length normalization, PLDA has
delivered state-of-the-art performance in speaker verification.
Letting $w_{ij}$ denote the i-vector of the $j^{th}$ utterance (session) of the $i^{th}$ speaker,
the PLDA model can be formulated as follows:

\[
w_{ij}  =  u + V y_{i} + z_{ij},
\]

\noindent where $u$ is the speaker-independent global factor, $y_{i}$ and $z_{ij}$ represent the speaker-level
and utterance-level factors, respectively. The matrix $V$ consists of the basis of the speaker subspace. Note that
both $y_i$ and $z_{ij}$ are assumed to follow a diagonal full-rank Gaussian prior.
The model can be trained via an EM algorithm~\cite{prince2007probabilistic},
and the similarity of two i-vectors can be computed as the ratio of the evidence (likelihood) of
two hypothesises: whether or not the two i-vectors belong to the same speaker~\cite{Yang2012plda}.

\subsection{Knowledge-based weak training}

\begin{figure*}[htb]
\centering
\includegraphics[width=0.9\linewidth]{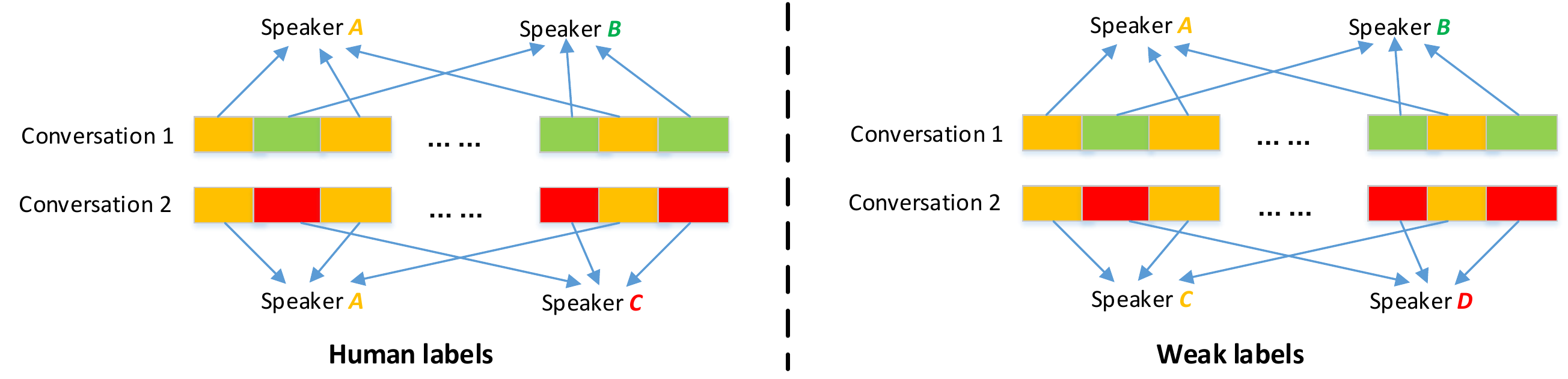}
\caption{Illustration of the difference between human labels and weak labels.}
\label{fig:labels}
\end{figure*}

We propose a weak training approach that relies on some prior knowledge to get cheap
labels for unlabelled data. For example in the customer service domain that the
paper focuses on, we utilize two pieces of prior
knowledge: (1) there are only a few (often two)
participants in a single session, and they can be easily separated; (2) the speakers in different
sessions are probably different, especially for customers. By these knowledge, an utterance can
be simply assigned a label that involves a session ID and a local speaker ID, i.e., an ID is valid only within the session.
These labels are not fully correct (so are weak labels),
but in most cases they are.

Figure~\ref{fig:labels} illustrates the difference between human labels and
the weak labels derived from the above prior knowledge,
where each speaker is represented by a particular color.
For human labels, the segments from the same speaker but different sessions
are correctly labelled. For weak labels, speakers in different sessions
are labelled as different, even if they are actually the same. Once the weak labels
are generated, the PLDA training is conducted as usual as with human labels.

\subsection{Relation to other methods}

The knowledge-based weak training proposed here is related to the semi-supervised PLDA training in ~\cite{garcia2014improving}.
Both of them rely on weak and cheap labels,
but the labels are produced in different ways: the knowledge-based weak training relies on
domain-specific prior knowledge, and its performance is determined by the
correctness of the knowledge; the semi-supervised training relies on
the existing PLDA model, and the performance is determined by the quality of the
existing model. From this perspective, the semi-supervised training can be regarded as a
model-based weak training. We argue that the knowledge-based weak training is
superior in scenarios where human-labelled data are insufficient that a strong
primary PLDA is not available.

The knowledge-based weak training is also related to unsupervised PLDA adaptation~\cite{garcia2014improving,villalba2014unsupervised,liu2014utilization,wang2016domain,rahman2015dataset}.
Both methods make use of the distribution information of
unlabelled data and thus can be employed to perform model adaptation.
The difference is that the weak training also utilizes speaker-discriminant
information, which, although noisy, is still beneficial if the knowledge is
mostly correct. We therefore conjecture that the knowledge-based weak training
is more effective than unsupervised adaptation in scenarios where the discriminative
information is desirable.

\section{Experiment}
\label{sec:exp}

The proposed weak training approach is tested on a practical speaker verification system
trained with a large-scale telephony customer service archive. The system is implemented
based on the GMM-ivector framework. We first present the data profile and then report the results.

\subsection{Data and configurations}

The training data used to train the GMM-ivector system are composed of $500$ hours
of conversational speech signals sampled from a large-scale telephony customer service archive.
These data are used to train the
UBMs and the T matrix of the i-vector model.
The development data used to train the PLDA model are divided into two data sets:
the STRONG set and the WEAK set that are labelled by human and the prior knowledge
described in the previous section, respectively.
Note that, the acoustic condition of the WEAK set is more close
to that of the evaluation data, which means that the WEAK set can be
regarded as in-domain and therefore any improvement with this set
could be partially due to model adaptation.

The STRONG set involves speech signals of $2,000$ speakers, and
the WEAK set consists of $2,000$ double-channel sessions, each
with two speakers. Each session consists of a customer channel and
a service channel, and the two channels are separated physically.
The WEAK set of the customer channel forms a WEAK-customer subset
and the WEAK set of the service channel forms a WEAK-service subset.
We distinguish customer data and service data because they hold very different
properties, particularly the probability that
the `different session, different speaker' assumption holds.
Finally, we sample $1,000$ sessions from WEAK-customer
and $1,000$ sessions from WEAK-service, composing a WEAK-mix subset.
More details about the development data are shown in Table~\ref{tab:dev}.

\begin{table}[htb]
\normalsize
\caption{Development set for PLDA training.}
\label{tab:dev}
\centering
\begin{tabular}{l|c|c}
\hline
                                &    $\#$ of Spks/Sessions &  $\#$ of Utts     \\
\hline
\hline
    \textbf{STRONG}             & 2,000         &    15,718           \\
\hline
    \textbf{WEAK-customer}      & 2,000         &    21,463           \\
\hline
    \textbf{WEAK-service}       & 2,000         &    25,852           \\
\hline
    \textbf{WEAK-mix}           & 2,000         &    23,987           \\
\hline

\end{tabular}
\end{table}

The evaluation set involves $1,236$ speakers and the enrollment speech for each speaker is $30$ seconds in length.
The length of the test utterances is $15$ seconds and each speaker contains about $6$ test utterances. By
pair-wised composition, $9,469,462$ trails are constructed, including $7,649$ target trials and $9,461,813$ imposter trails.

The acoustic feature used in our experiments is the $60$-dimensional Mel frequency cepstral coefficients (MFCCs),
which involves $20$-dimensional static components plus the first and second order derivatives.
The frame size is $20$ ms and the frame shift is $10$ ms. The UBM
involves $1,024$ Gaussian components and the dimensionality of the i-vector space is $400$.
The performance is evaluated in terms of Equal Error Rate (EER)~\cite{greenberg20132012}.

\subsection{Strong and weak training}
\label{sec:basic}

The first experiment studies the performance of the knowledge-based weak PLDA training,
and compare it with the strong training that uses the human-labelled data.
The EER results are shown in Table~\ref{tab:eer}, where the results with the STRONG set
and three WEAK subsets are reported. For comparison, the results with
cosine scoring (NO PLDA) are also presented. We first observe that most of the PLDA
models outperform the cosine scoring. This is particular interesting for the weak training
approach, where only inaccurate labels are used. This confirms our conjecture that
it is possible to use weak labels derived from prior knowledge to
train PLDA, at least in scenarios where the prior knowledge is correct.

\begin{table}[htb]
\normalsize
\caption{EER(\%) results of strong and weak training. }
\label{tab:eer}
\centering
\begin{tabular}{l|c}
\hline
Scoring Method    & EER\% \\
\hline
Cosine              & 2.88 \\
\hline
PLDA: STRONG        & 2.25 \\
\hline
PLDA: WEAK-customer & 2.47 \\
PLDA: WEAK-service  & 2.94 \\
PLDA: WEAK-mix      & 2.55 \\
\hline
\end{tabular}
\end{table}

Comparing the results with the three WEAK subsets,
it can be observed that WEAK-customer delivers the best performance,
while WEAK-service shows the worst (the performance is actually worse than with Cosine scoring).
This is also understandable, since
the number of service people is limited (about $200$) so speaker labels
of the $2,000$ sessions of the WEAK-service subset are probably incorrect
(sessions of the same speaker are labelled as distinct speaker IDs).
In contrast, the probability that
two customers appear in the $2,000$ sessions in WEAK-customer is fairly low,
which means a perfect match between the prior knowledge and the real data,
leads to the good performance.

\begin{figure}[htb]
\centering
\includegraphics[width=0.9\linewidth]{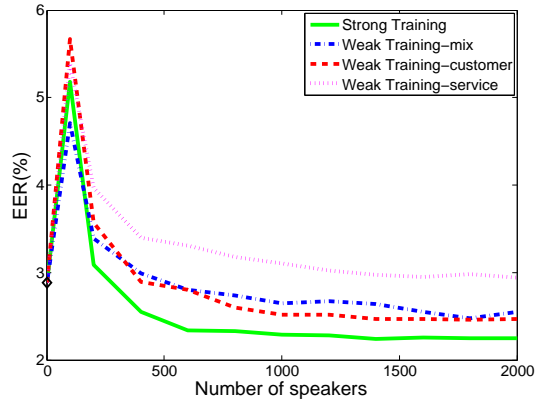}
\caption{Performance of strong and weak training with different amount of training data.}
\label{fig:strong-weak}
\end{figure}

In the second experiment, we investigate the performance of different PLDA training methods
with various amount of training data. The results are shown in Figure~\ref{fig:strong-weak},
where the data volume is controlled by the number of speakers.
The diamond at the starting point of each curve represents the performance with the cosine scoring.
It can be seen that with limited data (less than $200$), the PLDA models, despite strong training
or weak training applied, do not provide better performance than the simple cosine scoring.
With more data, the PLDA models offer better performance than the cosine baseline. The
strong training is superior to the weak training, and for the weak training, the model
trained with WEAK-customer shows better performance than with WEAK-service, due to the reason
that has been discussed already.

\subsection{Pooled training}

\begin{figure}[htb!]
\centering
\includegraphics[width=1\linewidth]{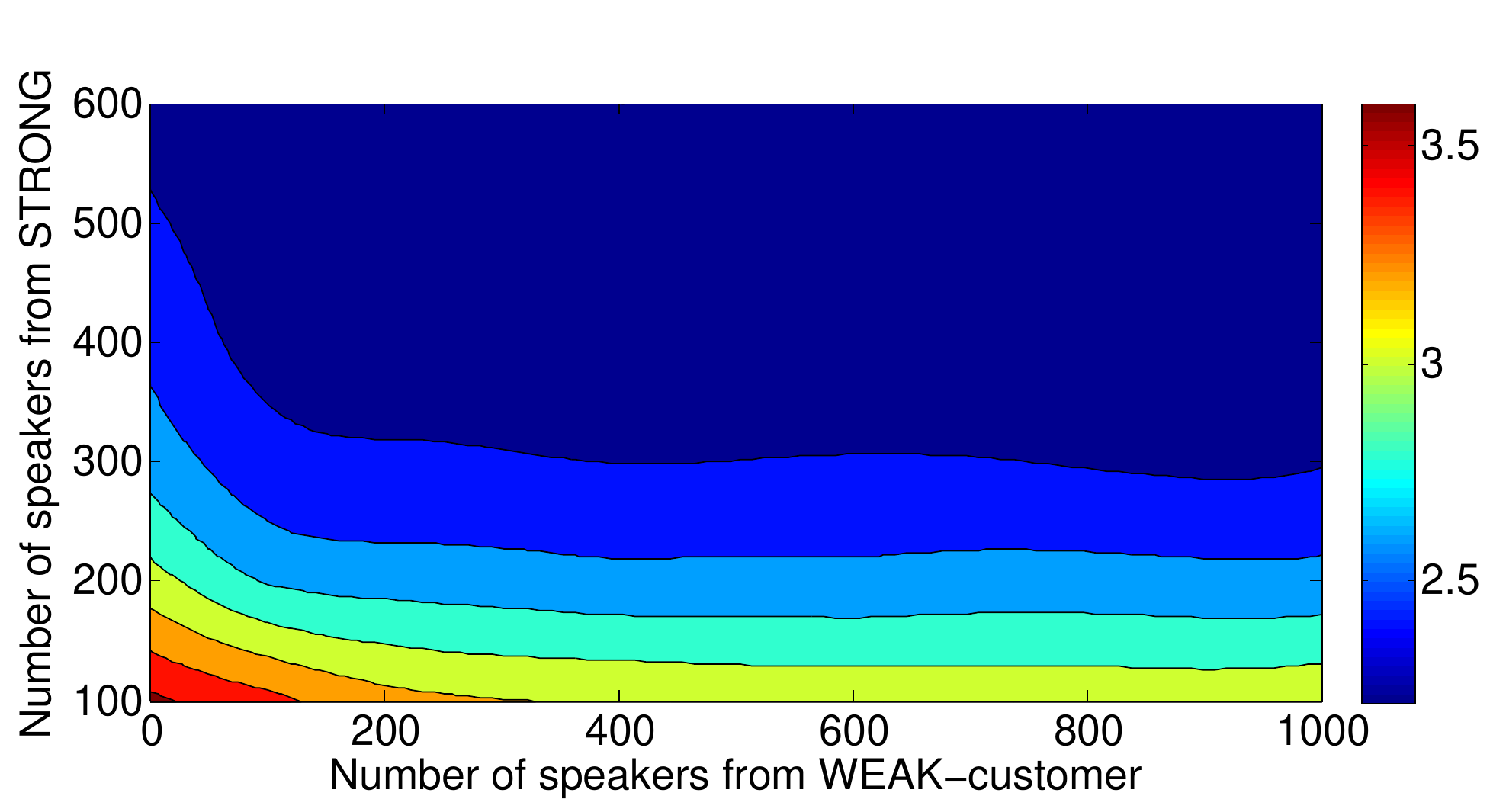}
\caption{Contour of EER results with pooled training, with various amount of data from
STRONG (y-axis) and WEAK-customer(x-axis).}
\label{fig:strong-weak-2}
\end{figure}

In this experiment, we assume limited human-labelled data and use weakly labelled data to
enhance the PLDA model. More precisely, the weakly-labelled data are augmented with the human-labelled
data to train the PLDA, which we call `pooled training'. According to the experience in the
last experiment, only the data in WEAK-customer are used for data augmentation.
Figure~\ref{fig:strong-weak-2} shows the
contour of the performance of the pooled training, with various amount of data from
STRONG and WEAK-customer. It can be seen that if the human-labelled data are limited (the number of
speakers is less than 500), augmenting weakly-labelled data offers clear performance improvement,
and the more data augmented, the more performance improved. However, the effectiveness of the
augmentation is not unlimited: the additional contribution becomes marginal if the amount of
weakly-labelled data is more than $800$ speakers. This is not surprising considering the noise involved
in the data. Figure~\ref{fig:strong-weak-2} also suggests an interesting concept of `substitution amount', i.e.,
how many weakly-labelled data can substitute for a certain amount of human-labelled data. The contour in
Figure~\ref{fig:strong-weak-2} indicates that the more human-labelled data are provided, the more difficult they can be
substituted by weakly-labelled data. In other words, the most value of human-labelled
data is to provide additional performance gains, instead of offering baseline performance.
This suggests an active learning approach that is under investigation.

\begin{figure}[htb]
\centering
\includegraphics[width=1\linewidth]{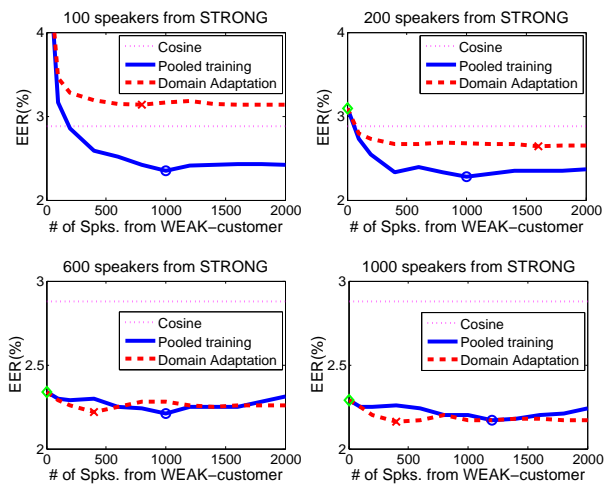}
\caption{Performance of pooled training and unsupervised adaptation.
The green diamonds represent the performance with strong training,
the blue circles represent the best performance of pooled training,
and the red crosses represent the best performance of unsupervised adaptation.
}
\label{fig:pool}
\end{figure}

Finally, we compare the pooled training and unsupervised learning. Note that the WEAK dataset is more close
to the evaluation set in the acoustic condition, so both methods play the role of model adaptation.
Figure~\ref{fig:pool} shows the results, where the four plots present configurations
with different amount of human-labelled data to train the initial PLDA model. Again, only
the WEAK-customer subset is used as the adaptation data.

From Figure~\ref{fig:pool}, it can be observed that if the human-labelled data are limited,
both the pooled training and the unsupervised adaptation offer clear performance improvement,
though the pooled training is more effective.
We attribute the superiority of the pooled training to the fact that it not only adapts to
the new acoustic condition, but also utilizes the speaker-related discriminant information
associated with the weak labels. Since the human-labelled data are limited,
the initial PLDA is not strong, and therefore the additional discriminant
information is essentially valuable, leading to the clear advantage with the pooled training.
If the human-labelled data are sufficient, the initial PLDA model covers most of the
acoustic conditions and holds sufficient discriminative capability,
diminishing the contribution of both pooled training and unsupervised adaptation.

\section{Conclusion}
\label{sec:conc}

This paper proposed a knowledge-based weak training approach for PLDA and verified
its potential in speaker verification.
Based on the assumption that speakers in different sessions are different,
weak labels can be easily produced and used as supplemental data to
train PLDA.
Our experiments on a large-scale customer service archive demonstrated
that the weak training approach works well when the `different session, different speaker'
assumption is held. This approach is most effective when human-labelled data are limited,
even outperforming the unsupervised adaptation method.
Future work will investigate the possibility to utilize both the knowledge-based weak labels and
model-based weak labels, and investigate active learning to select the most valuable
data for human labeling.



\bibliographystyle{IEEEtran}

\bibliography{refs}

\begin{thebibliography}{10}
\providecommand{\url}[1]{#1}
\csname url@samestyle\endcsname
\providecommand{\newblock}{\relax}
\providecommand{\bibinfo}[2]{#2}
\providecommand{\BIBentrySTDinterwordspacing}{\spaceskip=0pt\relax}
\providecommand{\BIBentryALTinterwordstretchfactor}{4}
\providecommand{\BIBentryALTinterwordspacing}{\spaceskip=\fontdimen2\font plus
\BIBentryALTinterwordstretchfactor\fontdimen3\font minus
  \fontdimen4\font\relax}
\providecommand{\BIBforeignlanguage}[2]{{%
\expandafter\ifx\csname l@#1\endcsname\relax
\typeout{** WARNING: IEEEtran.bst: No hyphenation pattern has been}%
\typeout{** loaded for the language `#1'. Using the pattern for}%
\typeout{** the default language instead.}%
\else
\language=\csname l@#1\endcsname
\fi
#2}}
\providecommand{\BIBdecl}{\relax}
\BIBdecl

\bibitem{dehak2011front}
N.~Dehak, P.~Kenny, R.~Dehak, P.~Dumouchel, and P.~Ouellet, ``Front-end factor
  analysis for speaker verification,'' \emph{IEEE Transactions on Audio,
  Speech, and Language Processing}, vol.~19, no.~4, pp. 788--798, 2011.

\bibitem{garcia2011analysis}
D.~Garcia-Romero and C.~Y. Espy-Wilson, ``Analysis of i-vector length
  normalization in speaker recognition systems,'' in \emph{Proceedings of the
  Annual Conference of International Speech Communication Association
  (INTERSPEECH)}, 2011, pp. 249--252.

\bibitem{lei2014novel}
Y.~Lei, N.~Scheffer, L.~Ferrer, and M.~McLaren, ``A novel scheme for speaker
  recognition using a phonetically-aware deep neural network,'' in
  \emph{Proceedings of IEEE International Conference on Acoustics, Speech and
  Signal Processing (ICASSP)}.\hskip 1em plus 0.5em minus 0.4em\relax IEEE,
  2014, pp. 1695--1699.

\bibitem{kenny2010bayesian}
P.~Kenny, ``Bayesian speaker verification with heavy-tailed priors,'' in
  \emph{Proceedings of Odyssey}, 2010.

\bibitem{fisher2004}
C.~Cieri, D.~Miller, and K.~Walker, ``The fisher corpus:a resource for the next
  generations of speech-to-text,'' in \emph{The Fourth International Conference
  on Language Resources and Evaluation (LREC),2004}, 2004, pp. 69--71.

\bibitem{switch}
``The switchboard-1 telephone speech corpus introduction.''\hskip 1em plus
  0.5em minus 0.4em\relax https://catalog.ldc.upenn.edu/LDC97S62.

\bibitem{wang2016domain}
Q.~Wang, H.~Yamamoto, and T.~Koshinaka, ``Domain adaptation using maximum
  likelihood linear transformation for plda-based speaker verification,'' in
  \emph{Proceedings of IEEE International Conference on Acoustics, Speech and
  Signal Processing (ICASSP)}.\hskip 1em plus 0.5em minus 0.4em\relax IEEE,
  2016, pp. 5110--5114.

\bibitem{rahman2015dataset}
M.~H. Rahman, A.~Kanagasundaram, D.~Dean, and S.~Sridharan, ``Dataset-invariant
  covariance normalization for out-domain plda speaker verification,'' in
  \emph{Proceedings of the Annual Conference of International Speech
  Communication Association (INTERSPEECH)}, 2015, pp. 1017--1021.

\bibitem{garcia2014improving}
D.~Garcia-Romero, X.~Zhang, A.~McCree, and D.~Povey, ``Improving speaker
  recognition performance in the domain adaptation challenge using deep neural
  networks,'' in \emph{Spoken Language Technology Workshop (SLT)}.\hskip 1em
  plus 0.5em minus 0.4em\relax IEEE, 2014, pp. 378--383.

\bibitem{villalba2014unsupervised}
J.~Villalba and E.~Lleida, ``Unsupervised adaptation of plda by using
  variational bayes methods,'' in \emph{Proceedings of IEEE International
  Conference on Acoustics, Speech and Signal Processing (ICASSP)}.\hskip 1em
  plus 0.5em minus 0.4em\relax IEEE, 2014, pp. 744--748.

\bibitem{liu2014utilization}
G.~Liu, C.~Yu, N.~Shokouhi, A.~Misra, H.~Xing, and J.~H. Hansen, ``Utilization
  of unlabeled development data for speaker verification,'' in \emph{Spoken
  Language Technology Workshop (SLT)}.\hskip 1em plus 0.5em minus 0.4em\relax
  IEEE, 2014, pp. 418--423.

\bibitem{prince2007probabilistic}
S.~J. Prince and J.~H. Elder, ``Probabilistic linear discriminant analysis for
  inferences about identity,'' in \emph{Proceedings of 11th International
  Conference on Computer Vision (ICCV)}.\hskip 1em plus 0.5em minus 0.4em\relax
  IEEE, 2007, pp. 1--8.

\bibitem{Yang2012plda}
Y.~Hai, L.~Yan, and X.~Fei, ``Sparse probabilistic linear discriminant analysis
  for speaker verification,'' in \emph{Proceedings of the Annual Conference of
  International Speech Communication Association (INTERSPEECH)}, 2012, pp.
  2658--2661.

\bibitem{greenberg20132012}
C.~S. Greenberg, V.~M. Stanford, A.~F. Martin, M.~Yadagiri, G.~R. Doddington,
  J.~J. Godfrey, and J.~Hernandez-Cordero, ``The 2012 nist speaker recognition
  evaluation,'' in \emph{Proceedings of the Annual Conference of International
  Speech Communication Association (INTERSPEECH)}, 2013, pp. 1971--1975.

\end{thebibliography}

\end{document}